\documentclass[runningheads]{llncs}

\usepackage{graphicx}
\usepackage{booktabs}
\usepackage{comment}

\usepackage[accsupp]{axessibility}  

\usepackage[pagebackref,breaklinks,colorlinks,citecolor=blue]{hyperref}

\usepackage{orcidlink}
\newcommand{\inlineeqnum}{\refstepcounter{equation}~~\mbox{(\theequation)}}

\begin{document}

\title{ Camera Calibration and Stereo via a Single Image of a Spherical Mirror} 


\author{Nissim Barzilay \& Ofek Narinsky \& Michael Werman}


\institute{Department of Computer Science, The Hebrew University of Jerusalem}

\maketitle

\begin{abstract}
This paper presents a novel technique for camera calibration using a single view that incorporates a spherical mirror. 
Leveraging the distinct characteristics of the sphere's contour visible in the image and its reflections, we showcase the effectiveness of our method in achieving precise calibration. 
 Furthermore, the reflection from the mirrored surface provides additional information about the surrounding scene beyond the image frame.

Our method paves the way for the development of simple catadioptric stereo systems. We explore the challenges and opportunities associated with employing a single mirrored sphere, highlighting the potential applications of this setup in practical scenarios. The paper delves into the intricacies of the geometry and calibration procedures involved in catadioptric stereo utilizing a spherical mirror.

Experimental results, encompassing both synthetic and real-world data, are presented to illustrate the feasibility and accuracy of our approach.
\end{abstract}

\section{Introduction}
\label{sec:intro}
\begin{figure}[ht!]
    \centering
    \includegraphics[width=0.45\columnwidth]{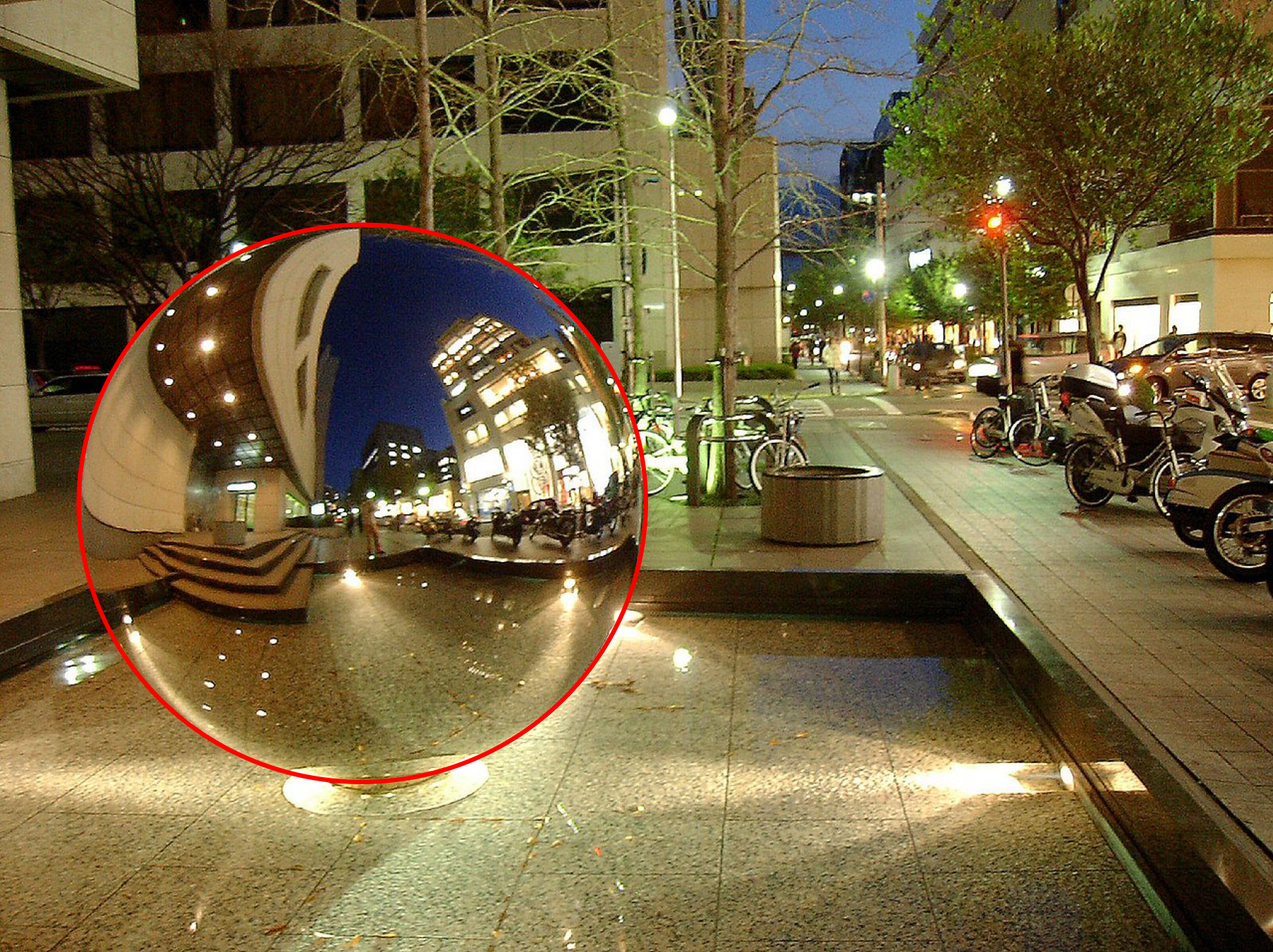}
    \caption{Spherical mirror in scene}
    \label{fig:find-c}
\end{figure}

Incorporating  spherical mirrors in a catadioptric 
imaging system makes it possible to observe a wide area with a
relatively small mirror.
 Research and analysis of  catadioptric systems based on spherical mirrors can be found in various papers
\cite{s18020408,BARONE201883,surv}.

Inspired by the concepts introduced in  \cite{710852,9621989}, which utilized two spheres in the camera's field of view for obtaining stereo information, our focus is on the more practical scenario of employing a single mirrored sphere. Our research aims to present a method capitalizing on the unique attributes of a single mirrored sphere for both camera matrix calibration and catadioptric stereo.

Our approach only requires the image to show  part of the sphere's contour and one of the following;  the reflection of the camera, two pairs of corresponding points on and off the  spherical mirror, or a single correspondence in special cases.

This research extends to the practical implementation of a real-time system, showcasing the feasibility and efficacy of employing mirrors for stereo imaging as a compelling alternative to the established two-camera stereo methodologies. It is also applicable in scenarios where an accidental spherical mirror is present in the scene.

\section{Related Work}

Agrawal et al. \cite{1238428} and Zhang et al. \cite{4069265} presented a comprehensive method addressing both intrinsic and extrinsic calibration by strategically placing spheres at three or more locations.
For extrinsic calibration, these methods initially estimate the 3D positions of sphere centers in each camera coordinate system using known intrinsic parameters and the ellipses derived from the spheres' projection in the image.
The relative rotation and translation between two cameras are then determined through 3D point registration of the estimated sphere centers.

\section{Method}
\label{sec:pagestyle}

In this paper, we  assume:
\begin{itemize}
    \item A projective camera with no skew.
    \item The image contains a spherical mirror.
    \item The extrinsic parameters of the camera are
   
\[\begin{bmatrix}
1 & 0 & 0 & 0 \\
0 & 1 & 0 & 0 \\
0 & 0 & 1 & 0
\end{bmatrix}.\]
    \item The  unit is defined by the sphere's radius.
\end{itemize}

To 
calibrate the camera
we need to find the  sphere's contour and center in the image.
The sphere projects to an ellipse in the image \cite{min}. Let the conic be $v^TCv=0$, where $T$ denotes transposition, with $v$  the homogeneous coordinates of a point on the conic, and $C$ is the $3\times 3$ symmetric matrix (see Figure. \ref{fig:find-c}).

Locating  ellipses  in images is   a long-studied challenge, with various methods proposed to tackle it, including both traditional and deep learning approaches, for example,
\cite{cicconet2014ellipses,electronics12163431,fitzgibbon1999direct,lu2019arc,dong2021ellipse}.

\begin{figure}[ht!]
    \centering
    \includegraphics[width=0.96\columnwidth]{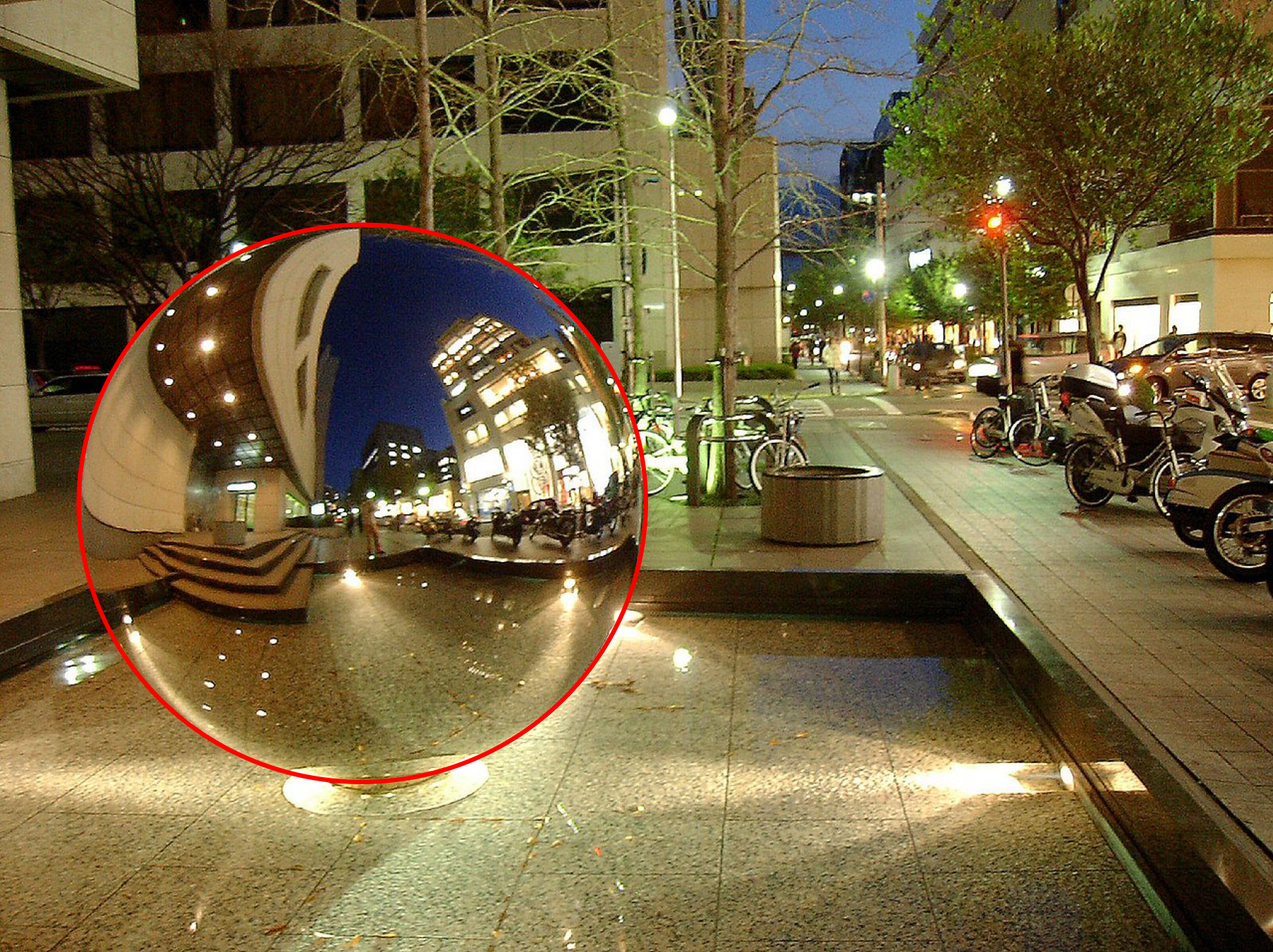}
    \caption{The ellipse which is the projected sphere's contour in red}
    \label{fig:find-c}
\end{figure}

\begin{figure}[ht!]
    \centering
    \includegraphics[width=0.80\columnwidth]{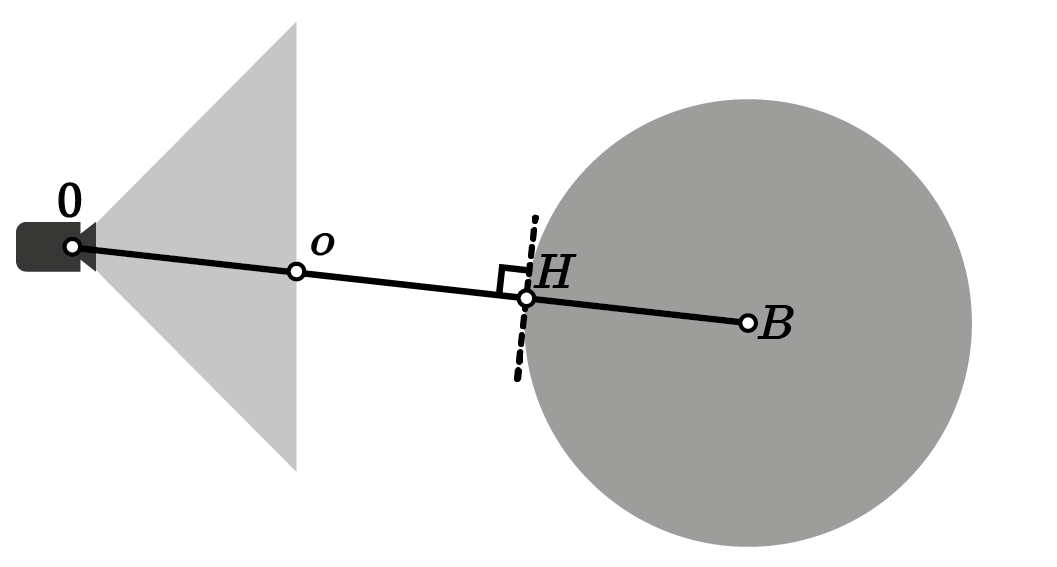}
    \caption{The rays from the camera to the mirror \(0\rightarrow H\), from the mirror to the camera \(H\rightarrow 0\) and the normal at the mirror coincide.}
    \label{fig:self-reflection}
    \includegraphics[width=0.90\columnwidth]{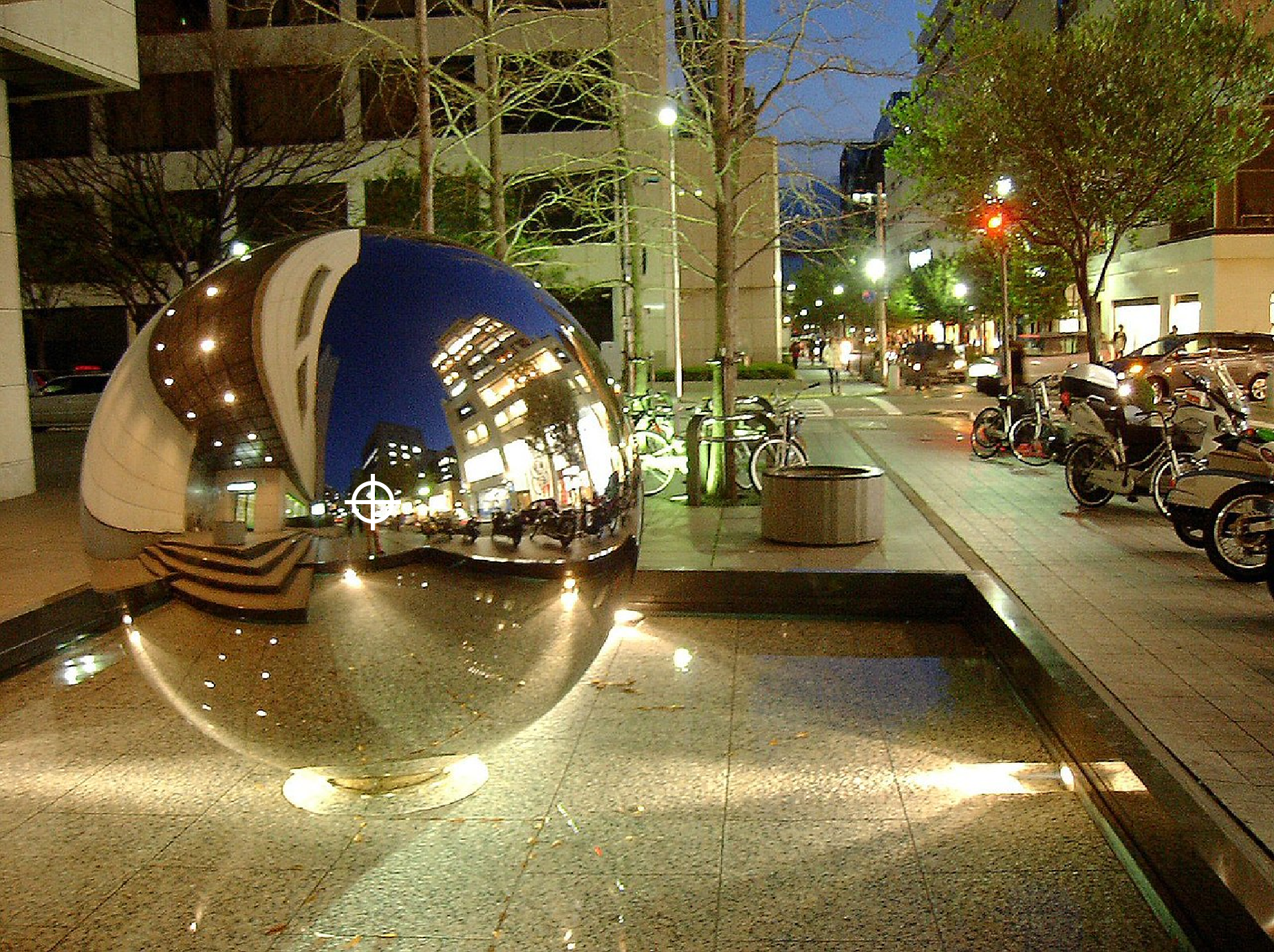}
    \caption{The center of the sphere in the image is at the camera's reflection.}
    \label{fig:find-o-1}
\end{figure}

\begin{figure}[ht!]
    \centering
    \includegraphics[width=0.96\columnwidth]{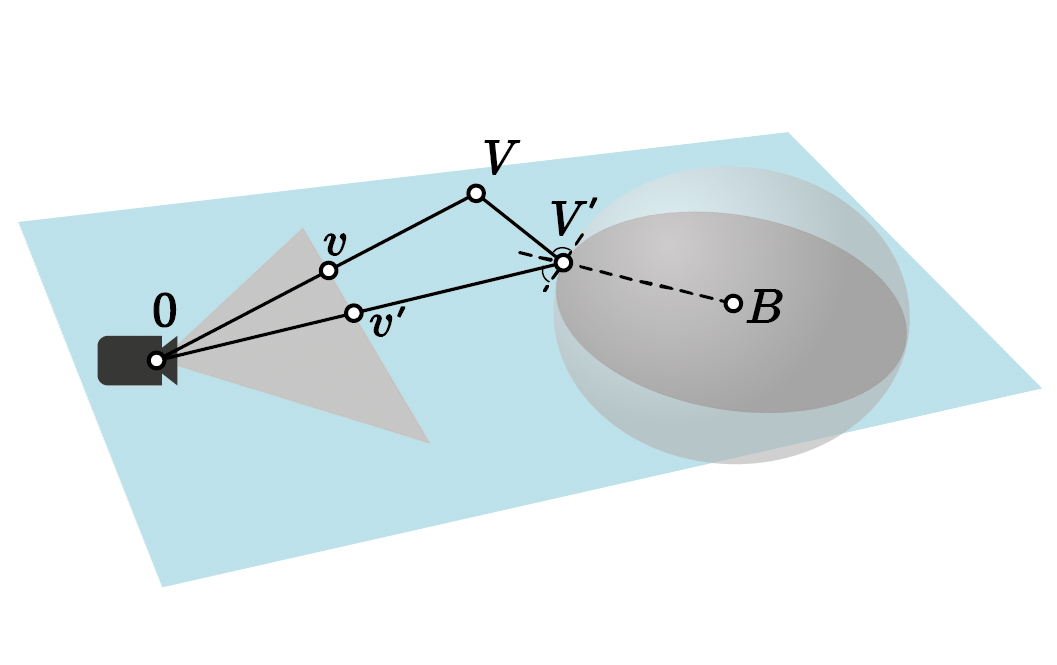}
    \caption{A 2d cross section of a pair of correspondence points}
    \label{fig:geometry-correspondence}
    \includegraphics[width=0.90\columnwidth]{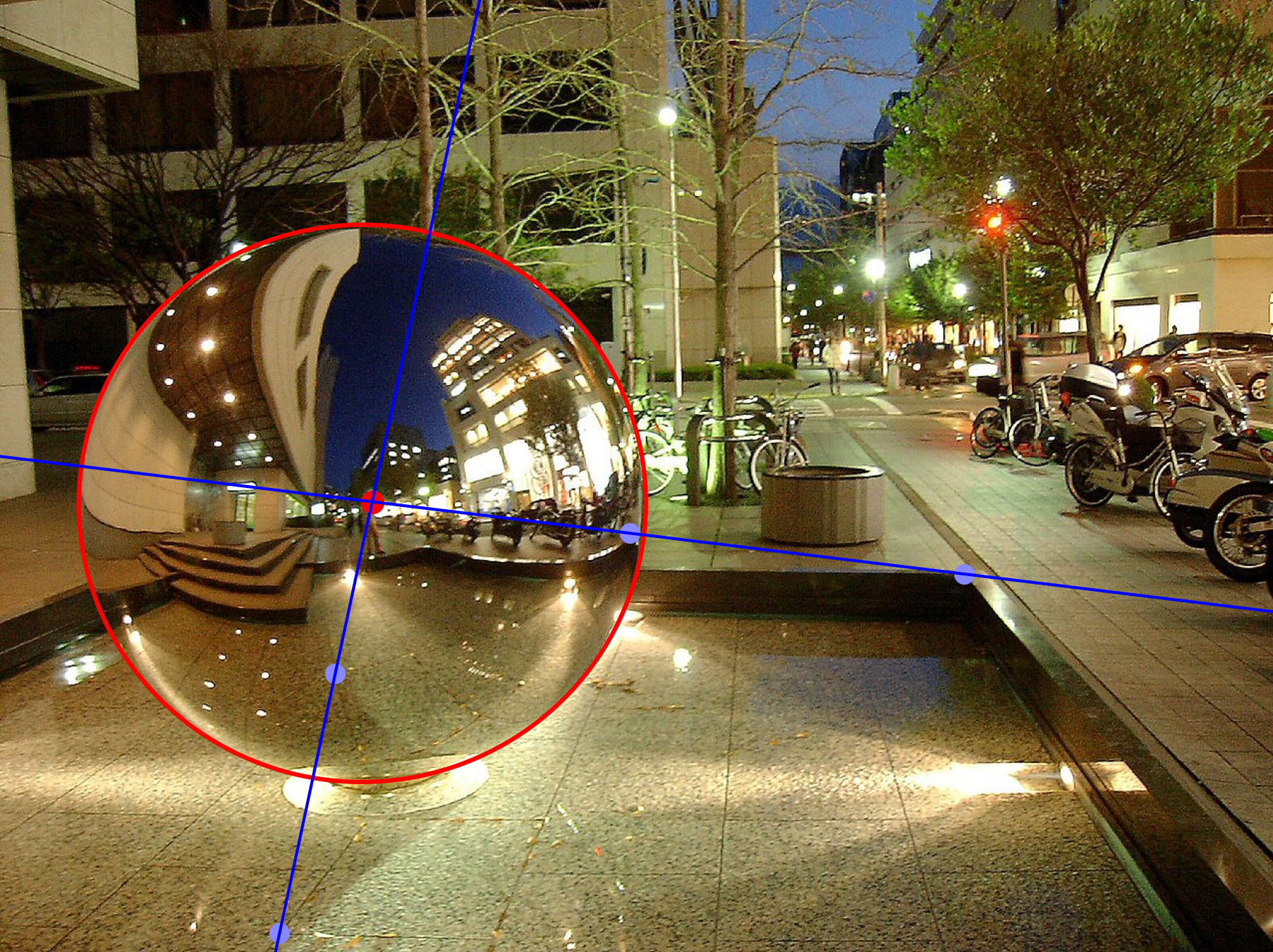}
    \caption{Finding the sphere's center in the image from two pairs of corresponding points.}
    \label{fig:find-o-2}
\end{figure}

\begin{figure}[ht!]
    \centering
    \includegraphics[width=0.90\columnwidth]{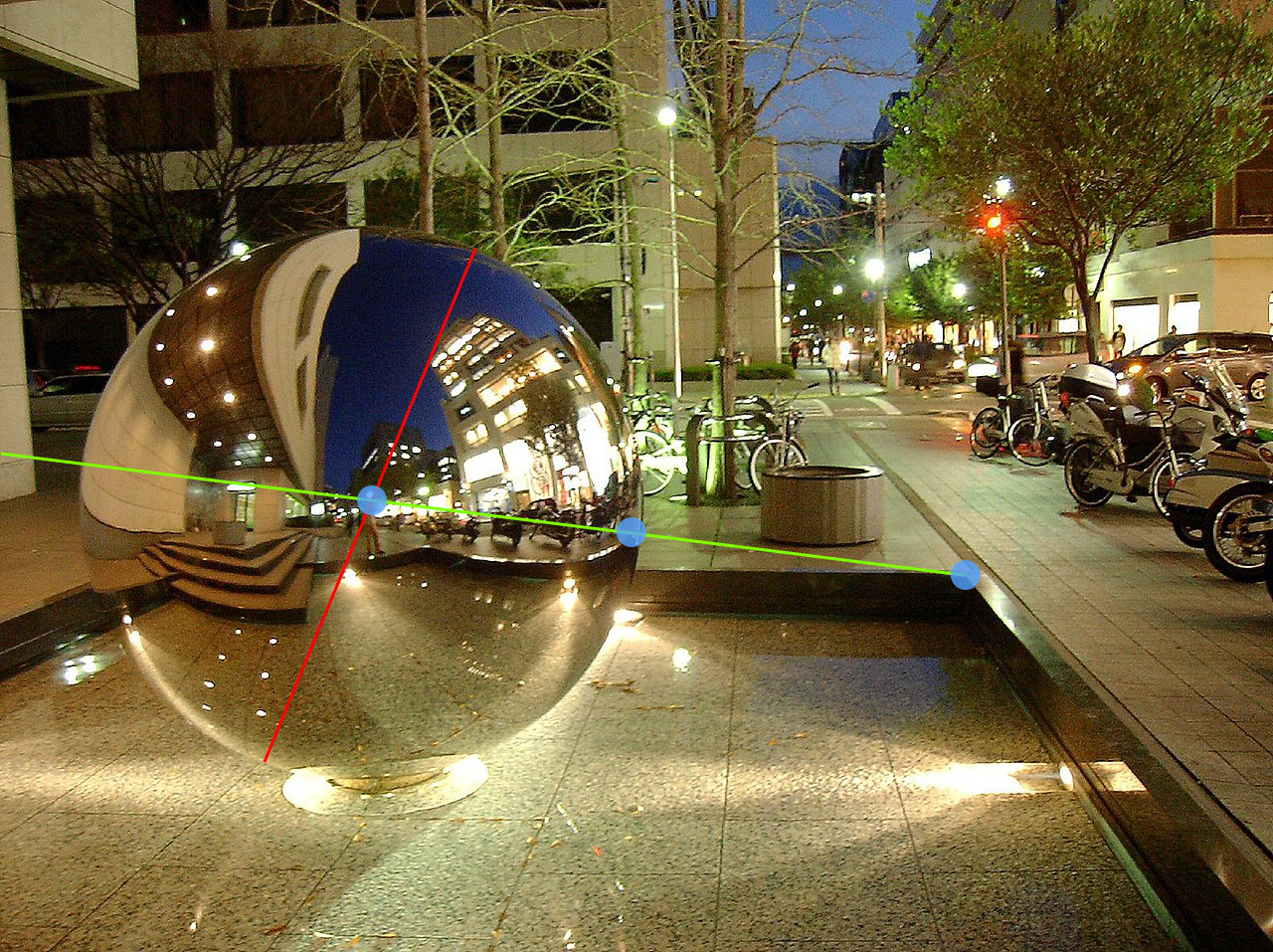}
    \caption{Finding the sphere's center from a single pair of corresponding points and the major axis
of the ellipse.
The green line  connects the corresponding points, while the red line represents the major axis of the ellipse.}
    \label{fig:find-o-3}
\end{figure}

Next, we find 
the  sphere's center in the image 
(\(O=\begin{bmatrix}o_x & o_y & 1\end{bmatrix}^T\)). \(O\),
can be determined by either of the following three methods: 
\begin{enumerate}
\item Locating the camera's reflection in the mirror (see Figure \ref{fig:self-reflection}, Figure \ref{fig:find-o-1}).
The rays from the camera to the mirror, from the mirror to the camera and the normal at the mirror coincide, thus the ray from the camera to its reflection in the mirror intersects the center of the sphere. So the image of the camera center is also the location of the {\it image} of the sphere's center.

\item Using 2 or more pairs of correspondence points, (see figure 
\ref{fig:geometry-correspondence}, figure \ref{fig:find-o-2}). 
Let \(v\) be the image of a 3D point \(V\) and \(v'\) the image of \(V\)'s reflection at \(V'\) then the rays from the camera to \(V'\), from \(V'\) to \(V\) and from \(V'\) to the sphere's center \(B\) (the normal) are coplanar and include the camera center thus project to the line in the image coincident to the sphere's center. The intersection of lines spanning corresponding points, on and off the mirror, is thus the image of the sphere's center.
    \item 
    If we assume that the  camera  has equal focal lengths,  \(f_x=f_y\).
    intersecting  the line containing a single pair of corresponding points and  the major axis of the ellipse, (see figure. \ref{fig:find-o-3}) suffices.
This follows from 
    the {\it axial} constraint \cite{min} which is the observation that the camera center,
    the sphere center, and  the major ellipse axis  are co-planar. Thus, the image of the sphere center is on the ellipse's major axis.
\end{enumerate}

We want to compute  the camera matrix \(P_ {3\times4}\) and the sphere's center  
\(B=\begin{bmatrix} b_x & b_y & b_z \end{bmatrix}^T\). We will use the radius of the sphere as the unit. Assuming a no skew camera 
\[P := \begin{bmatrix}
f_x & 0 & t_x & 0\\
0 & f_y & t_y & 0\\
0 & 0 & 1 & 0
\end{bmatrix}
=
\begin{bmatrix}
\begin{matrix}
    \mbox{\Huge K}
\end{matrix} & \begin{matrix} 0 \\ 0 \\ 0 \end{matrix}
\end{bmatrix}
\]
Where \(f_x,f_y\) are the focal lengths and \((t_x,t_y)\) is the principle point.

\begin{figure}[ht!]
    \centering
    \includegraphics[width=0.80\columnwidth]{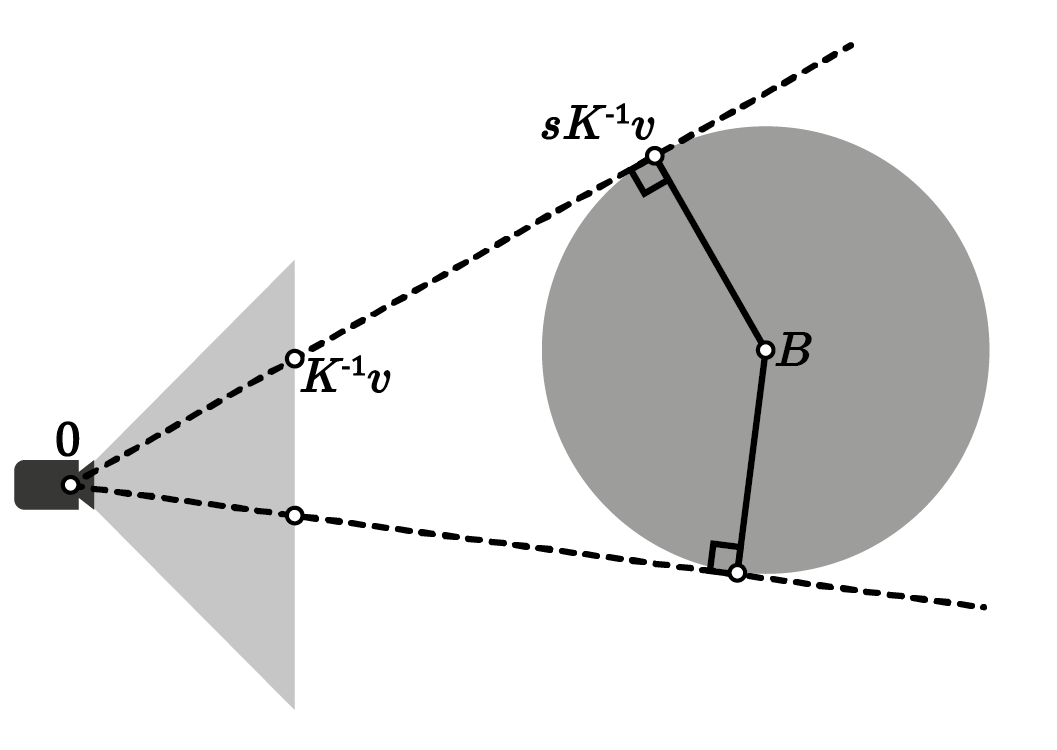}
    \caption{2D example of sphere outline. \(sK^{-1}V\) is perpendicular to \(B - sK^{-1}V\). The distance between \(sK^{-1}V\) and \(B\) is the radius, which is 1.}
    \label{fig:outline-2d}
\end{figure}

Let \(V=\begin{bmatrix}v_x&v_y&1\end{bmatrix}^T\in\mathbb{R}^3\) be a pixel on the projected contour of the sphere. Geometrically (see figure. \ref{fig:outline-2d}) this means that there is $s\in\mathbb{R}^+$ such that:
\begin{itemize}
    \item $\triangle 0,sK^{-1}V,B$ is a right  triangle.\\
    In other words $\langle sK^{-1}V,B-sK^{-1}v\rangle=0$.
    \item The distance between $sK^{-1}V$ and $B$ is the radius.\\
    The radius is our unit, so 
\(|sK^{-1}V-B|=1   \inlineeqnum \label{eqn:norm1} \).
\end{itemize}
We  simplify these equations to get:
$$
\langle K^{-1}V,B \rangle^2+(1-|B|^2)|K^{-1}V|^2=0$$
We  use the fact that an inner product can be represented by a matrix multiplication and   rewrite it as:
$$V^TK^{-T}(BB^T+(1-|B|^2)I)K^{-1}V=0$$
This is an equation of the conic section we already computed: $C$. Therefore, they are equivalent up to a scalar factor:
$$C=rK^{-T}(BB^T+(1-|B|^2)I)K^{-1}~~(r\in\mathbb{R})$$

We currently have 8 unknowns:
$$r,b_x,b_y,b_z,f_x,f_y,t_x,t_y$$
but equating the conic sections only gives 6 equations (Both matrices are symmetric). We first  get rid of $t_x,t_y$  by shifting the image so $(0,0)$ represents the center of the sphere.
We define:
$$S:=\begin{bmatrix}
1 & 0 & o_x \\
0 & 1 & o_y \\
0 & 0 & 1 \\
\end{bmatrix}$$
Since we know \(C,o\) we can compute the matrix $M:=S^TCS$
$$Q:=b_zK^{-1}S=\begin{bmatrix}
b_zf_x^{-1} & 0 & b_x \\
0 & b_zf_y^{-1} & b_y \\
0 & 0 & b_z \\
\end{bmatrix}$$
$$p=\frac{r}{b_z^2}$$
We get:
$$M=pQ^T(BB^T+(1-|B|^2)I)Q$$


Denote  
\(m_{ij}:=M[i,j]\)
The equation $M=pQ^T(BB^T+(1-|B|^2)I)Q$. can be expanded to a system of equations:
$$\left\{
\begin{aligned}
  & m_{11} = pf_x^{-2}b_z^2(b_x^2+1-|B|^2) \\
  & m_{22} = pf_y^{-2}b_z^2(b_y^2+1-|B|^2) \\
  & m_{33} = p|B|^2 \\
  & m_{12} = pf_x^{-1}f_y^{-1}b_xb_yb_z^2 \\
  & m_{13} = pf_x^{-1}b_xb_z \\
  & m_{23} = pf_y^{-1}b_yb_z \\
\end{aligned}
\right.$$
To solve these equations, First calculate $p$ and $|B|^2$:
$$
p=\frac{m_{13}m_{23}}{m_{12}},~~
|B|^2=\frac{m_{33}}{p}$$
Now we can calculate $b_x^2,b_y^2,b_z^2$:
$$
b_x^2=\frac{1-|B|^2}{\frac{m_{11}}{m_{13}^2}p-1},~~
b_y^2=\frac{1-|B|^2}{\frac{m_{22}}{m_{23}^2}p-1},~~
b_z^2=|B|^2-b_x^2-b_y^2$$
The choice of either  the positive or negative  square root of $b_x^2,b_y^2$    doesn't matter and it will be compensated by positive or negative $f_x,f_y$. However, $b_z$ should be positive as the sphere is  in front of the camera.
Now we can determine the values of $f_x,f_y$:
$$
f_x=\frac{pb_xb_z}{m_{13}},~~
f_y=\frac{pb_yb_z}{m_{23}}
$$
Notice $KB$ is the position of the sphere's center in the image, so $KB=b_zo$. Therefore we can determine the values based on our previous calculations:
$$
t_x=o_x-f_x\frac{b_x}{b_z},~~
t_y=o_y-f_y\frac{b_y}{b_z}
$$

Note that knowing both the sphere's and camera parameters suffice to  reconstruct the \(3D\) positions of all pairs of corresponding points by intersecting the corresponding rays.

\section{Results}
In the real image $1600$x$1196$ (see Figure \ref{fig:find-c}) the estimated sphere origin is:
$$b_x=-0.76,~~b_y=0.13,~b_z=5.07$$
$$f_x=-1744,~~f_y=1732,~~t_x=722,~~t_y=583$$

\begin{figure}[ht!]
    \centering
    \includegraphics[width=0.8\columnwidth]{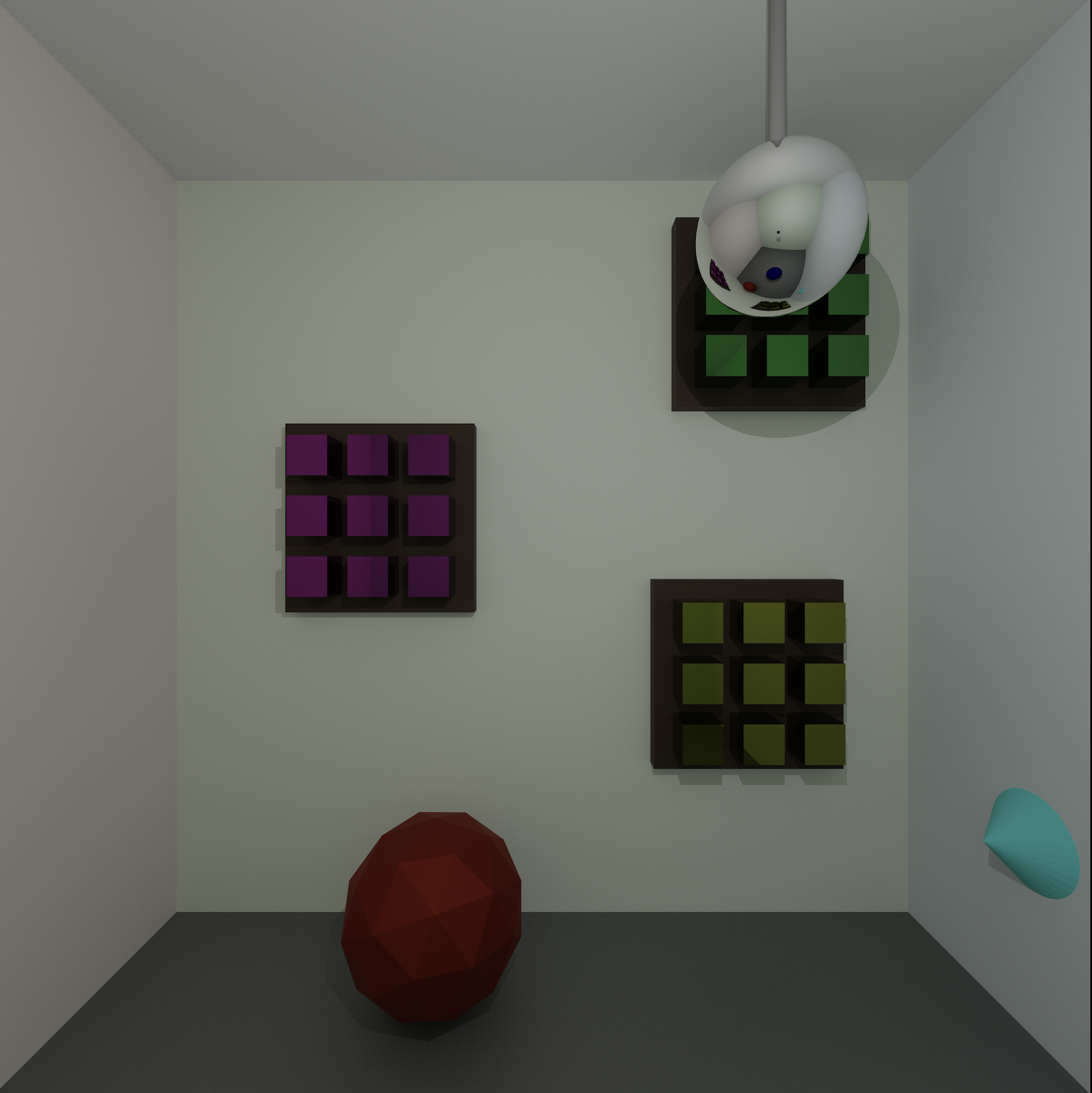}
    \caption{Synthetic Data  1}
    \label{fig:render}
\end{figure}

For example, we  tested our algorithm on a synthetic image of resolution 2048x2048, using the conic section and the reflection of the camera to calibrate the image (see figure. \ref{fig:render}):\\
First phase: We  selected  points on the sphere contour and calculated the conic.
Second phase: We estimate the center of the sphere in the image by locating the camera's reflection.
Now we apply our algorithm to calibrate the image.
Figure \ref{fig:render} 
real values:
$$b_x=3,~b_y=-4,~b_z=7$$
$$f_x=1024,~f_y=1024~~,~~t_x=1024,~t_y=1024$$
Our algorithm's result:
$$b_x=3.00,~~b_y=-3.94,~~b_z=7.03$$
$$f_x=1027.99,~f_y=1032.84,~~t_x=1024.34,~t_y=1016.94$$
the range of error is less than 1.5\%.

\begin{figure}[ht!]
    \centering
    \includegraphics[width=\columnwidth]{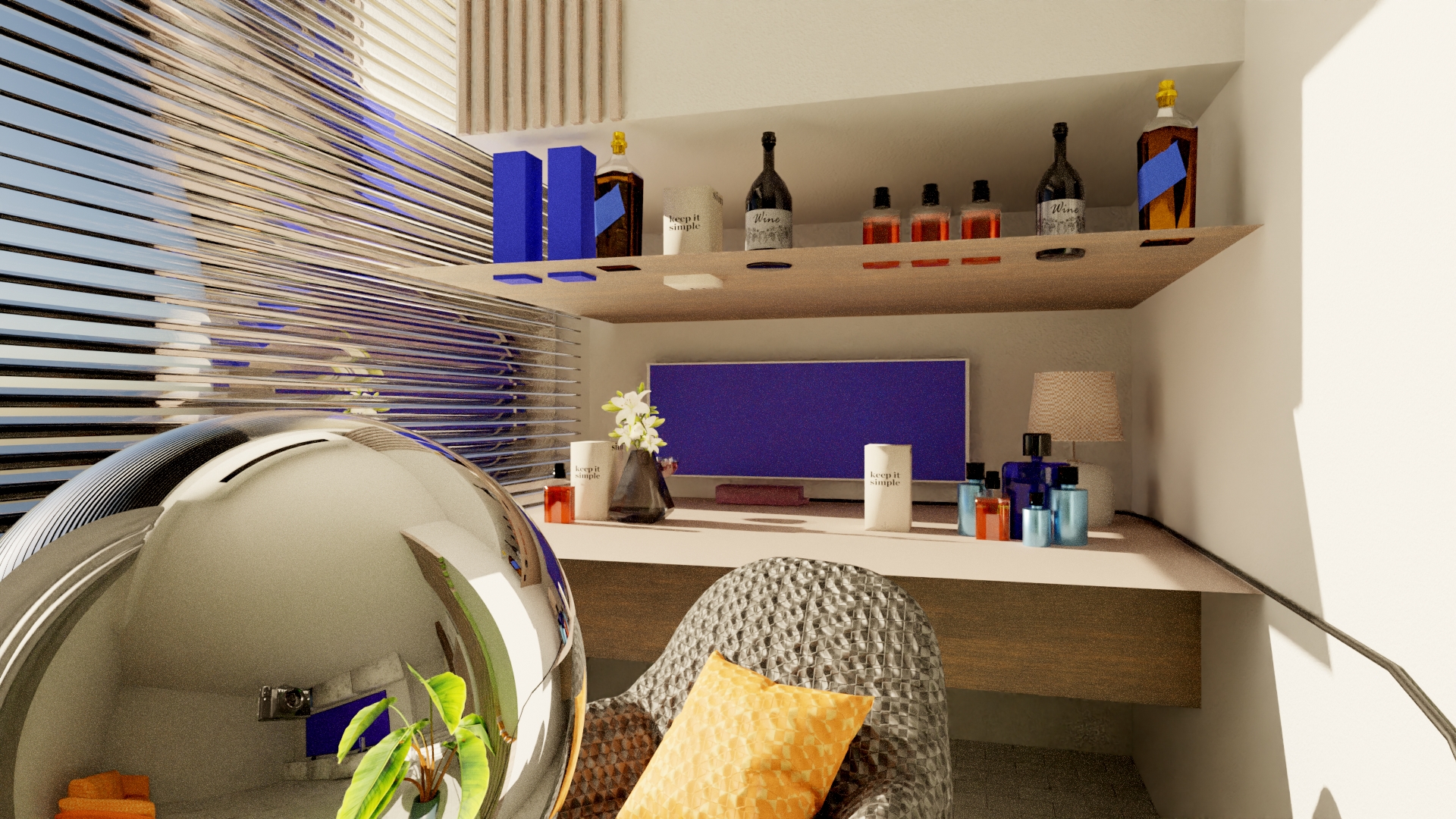}
    \caption{Synthetic Data  2}
    \label{fig:render_room_view}
\end{figure}

Figure \ref{fig:render_room_view} resolution $1920$x$1080$,
real values:
$$b_x=-1.5,~~b_y=3,~b_z=1$$
$$f_x=1144,~~f_y=1144~~,~~t_x=960,~~t_y=540$$
our algorithm's result:
$$b_x=-1.47,~~ b_y=3.07,~~ b_z=1$$
$$f_x=1179,~~ f_y=1167,~~ t_x=949, ~~t_y=535$$
the range of error is less than 3.1\%.

To verify our algorithm, we computed the length of  objects using two pairs of correspondence points (Figure 4) and a sphere with a radius of 5 cm. In Figure 10, we computed the height of the vase using two pairs of corresponding points.
We computed the ray for each point. Let \(v\),\(v'\), and \(u\), \(u'\) be pairs of correspondence points; we then calculate the rays in 3D space. This conversion involves scaling and translating the pixel coordinates. Next, we compute the 3D point representation where the ray intersects the correspondence point \(v'\), denoted as \(h\).
According to the equation we presented earlier, (\ref{eqn:norm1}), we define \(\text{offset} = h - B = sK^{-1}V-B \) with the condition \(|sK^{-1}V-B|=1\). The reflected vector is 
\[\text{reflect}= h - 2 * \langle \text{offset}, h\rangle * \text{offset}.\]
Given the reflected ray and the direct ray, we compute the 3D position of the point. The first and second phases are the same as described in the previous example.
$$b_x=1.30,~b_y=0.48,~b_z=5.62$$
$$f_x=2714,~f_y=2703~~,~~t_x=3052,~t_y=1664$$
The  height of the marker is 13cm,  computing the 3D points of \(v\), \(u\)  marked in red and their distance we  obtained is a height of 14 cm.
The real height of the tape dispenser is 5cm,  computing the 3D points of \(v\), \(u\) marked in blue and the distance we  calculated  a height of 5.05 cm.
\begin{figure}[ht!]
    \centering
    \includegraphics[width=\columnwidth]{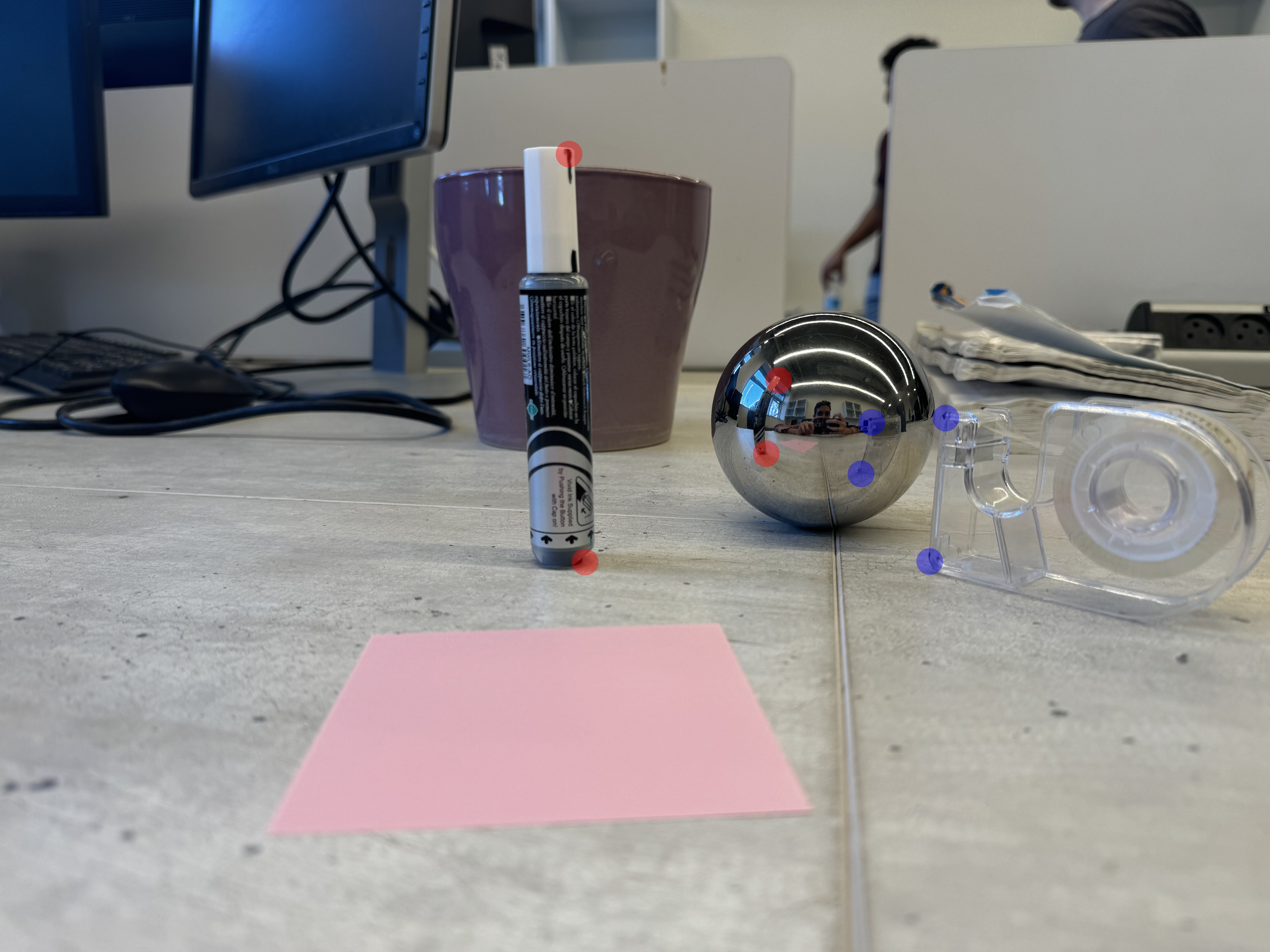}
    \caption{Height test}
    \label{fig:real_image}
\end{figure}
\section{Summary}
\label{sec:summary}
We presented a novel approach to calibrate the camera matrix using a single-view image. Our findings can help minimize the requirement for achieving this goal. Using our method, it's possible to  further analyse  the image, such as locating the 3d location of a point using a pair of corresponding points or creating an estimate for an omnidirectional image centered in the sphere's origin.

\bibliographystyle{splncs04}
\bibliography{main}
\end{document}